%% file: example.tex
\documentclass{article}

\usepackage[preprint]{corl_2026} 
\usepackage{multirow}
\usepackage{amsmath}
\usepackage{amssymb}
\usepackage{booktabs}
\usepackage{graphicx} 
\usepackage{subcaption}
\usepackage{graphicx}
\usepackage{makecell}
\usepackage[table]{xcolor}
\usepackage{caption}
\usepackage{enumitem}

\title{Envision4D: Envisioning Visual Futures via Feed-forward 4D Gaussian Splatting \\for Autonomous Driving}

\author{
  \vspace{-3pt}
  Qi Song$^1$, Yifei He$^1$, Chi Zhang$^2$, Zheng Fu$^1$,\\[2pt]
  \bfseries Xuhe Zhao$^1$, Mengmeng Yang$^1$, Kun Jiang$^1$, Rui Huang$^{2\dagger}$, Diange Yang$^{1\dagger}$\\[3pt]
  \normalfont $^1$Tsinghua University $\quad^2$The Chinese University of Hong Kong, Shenzhen\\[3pt]
  \texttt{https://maggiesong7.github.io/research/Envision4D/} \\
}

\begin{document}
\maketitle

\begin{center}
    \nopagebreak
    \vspace*{-26pt}
    \includegraphics[width=\textwidth]{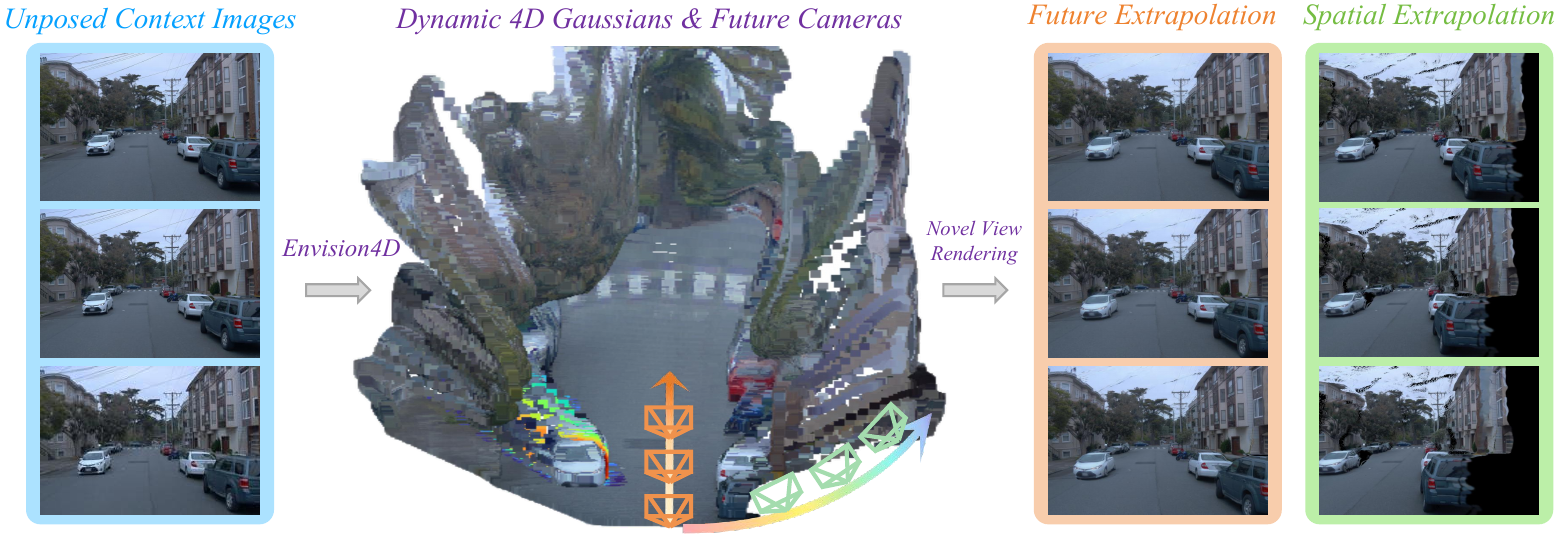}
    \captionof{figure}{\textbf{Illustration of Envision4D.} Envision4D reconstructs 4D Gaussians together with future poses in a self-supervised and feed-forward manner, enabling efficient dynamic scene extrapolation.}
    \label{fig:teaser}
\end{center}
\vspace{-6pt}

\begin{abstract}
Forecasting the future evolution of dynamic scenes is crucial in autonomous driving. However, existing feed-forward paradigms are primarily designed for interpolation. When extended to future extrapolation, they suffer from ghosting artifacts under large displacements and are constrained by simplified motion assumptions or strict future priors. To overcome these challenges, we propose Envision4D, a fully self-supervised feed-forward framework for pose-free future extrapolation. Specifically, we introduce a Future Pose Prediction module that infers future camera parameters via an iterative denoising process. Furthermore, to capture non-linear dynamics, we propose In-layer Temporal Attention and employ Conditioned Motion Lifting, which transforms the highly uncertain extrapolation process into robust relational mappings. Finally, a Progressive Training Strategy is utilized to stabilize unsupervised motion learning against error accumulation. Extensive experiments demonstrate that Envision4D achieves state-of-the-art performance, significantly outperforming existing methods in future view synthesis.
\end{abstract}



\section{Introduction}
Modeling dynamic scenes is fundamental to autonomous driving \citep{yan2025streetcrafter, song2024divide, yan2024street,chen2024omnire, chen2026periodic}, where forecasting future scene evolution is even more crucial for proactive decision-making \citep{wang2024omnidrive, fu2025orion, zhou2026opendrivevla}. Despite this imperative, current feed-forward paradigms \citep{hur2026ufo, sucar2026v, hu2025vggt4d, he2026dynamicvggt, yang2026neoverse} are primarily tailored for interpolating observed dynamics. When applied directly to future extrapolation, these interpolation-centric models fail to deal with the unbounded motion uncertainties, leading to severe error accumulation. 

The challenges lie in two main aspects:
(1) \textbf{Ineffective unsupervised motion modeling.} To fully unleash the potential of data-driven paradigms, methods like STORM \citep{yang2024storm} and Flux4D \citep{wang2025flux4d} explore unsupervised motion learning, bypassing the reliance on expensive explicit guidance such as optical flow \citep{lin2025movies, yang2026neoverse, he2026dynamicvggt}, off-the-shelf trackers \citep{fei2024driv3r, miao2026evolsplat4d,chen2025dggt}, or dynamic masks \citep{wu2025streamsplat,yu2026recondrive}. Unfortunately, without such strong priors, the unsupervised models struggle to handle large displacements, yielding severe ghosting artifacts around dynamic objects. 
(2) \textbf{Inflexible extrapolation mechanisms.} Existing methods are fundamentally constrained by simplified motion assumptions or strict future priors. For instance, restricted to single-step velocity estimation, NeoVerse \cite{yang2026neoverse} can only apply naive linear extrapolation, inevitably leading to trajectory deviations for complex dynamic objects. Meanwhile, 4DGT \citep{xu20254dgt} necessitates pre-given camera poses for future view synthesis, precluding true predictive forecasting. Consequently, these paradigms fail in open-world settings where future dynamics are highly non-linear and ego-trajectories remain completely unknown.

To address these issues, we propose Envision4D, a fully self-supervised feed-forward framework for dynamic scene extrapolation that operates on continuous images. As illustrated in Fig. \ref{fig:teaser}, Envision4D achieves robust future and spatial extrapolation in dynamic scenarios, successfully breaking free from the constraints of extra explicit guidance and restrictive future priors. 

Specifically, to eliminate the reliance on predefined ego-trajectories, we introduce a \textit{Future Pose Prediction} module that employs iterative denoising to adaptively infer future cameras. Built upon this, we rethink the velocity formulation by proposing time-conditioned motion, which is crucial for capturing real-world non-linear dynamics. Within this representation, \textit{In-layer Temporal Attention} is first employed to enhance the network's sensitivity to dynamic cues. Furthermore, our \textit{Conditioned Motion Lifting} mitigates extrapolation uncertainties by modeling source-to-target velocities conditioned on the current state, ego-motion, and temporal priors. Finally, a \textit{Progressive Training Strategy} stabilizes the unsupervised motion learning, allowing the model to gradually refine motion-aware features and prevent the severe error accumulation typical of unconstrained extrapolation.

Our main contributions are summarized as follows:
\begin{itemize}[leftmargin=*, itemsep=2pt, topsep=0pt]
    \item We propose Envision4D, a novel self-supervised 4DGS model capable of dynamic scene extrapolation in a future pose-free manner, without requiring any explicit motion guidance.
    
    \item We address the reliance on future ego-trajectories and linear motion assumptions by jointly inferring future camera poses and time-conditioned, non-linear motions. A tailored training strategy is further introduced to stabilize unsupervised motion learning.
    
    \item Extensive experiments demonstrate that Envision4D achieves state-of-the-art performance in future extrapolation and exhibits strong generalization capabilities in open-world driving scenarios.
\end{itemize}


\section{Related Work}
\label{sec:citations}

\textbf{Feed-forward Gaussian Splatting}
Recent years have witnessed a rapid transition of feed-forward Gaussian models from 3D \cite{chen2024mvsplat, yu2024mip, szymanowicz2024splatter, jiang2025anysplat,xu2025depthsplat} to 4D \cite{wu20244d, chen2026periodic, li2024st, lin2025movies} scene reconstruction. In the 3D domain, MVSplat \cite{chen2024mvsplat} leverages costvolume representations to enhance depth estimation, while ADGaussian \cite{song2025adgaussian} integrates multi-modal cues for robust geometry and visual modeling. To improve practical flexibility, pose-free frameworks like GGRt \cite{li2024ggrt} and NoPoSplat \cite{ye2024no} enable efficient static reconstruction directly from unposed images. Building upon these 3D foundations, recent 4D Gaussian models extend the paradigm by incorporating motion modeling. Specifically, DynamicVGGT \cite{he2026dynamicvggt} and NeoVerse \cite{yang2026neoverse} introduce motion attention modules on the top of VGGT \cite{wang2025vggt} backbone to capture inter-frame motions. ReconDrive \cite{yu2026recondrive} utilizes SAM2 \cite{ravi2024sam} for instance-level static-dynamic decomposition and computes object displacements via coordinate transformations, while DGGT \cite{chen2025dggt} employs external trackers for motion interpolation. Despite these advances, most feed-forward methods primarily focus on scene interpolation within observed frames, with limited exploration of extrapolation scenarios crucial for practical use. Moreover, they typically rely on pre-given future camera poses to perform extrapolation, failing to generalize to unseen open-world scenarios.

\textbf{Unsupervised Dynamic Reconstruction}
Reconstructing dynamic scenes without extra supervision, such as dynamic masks \cite{wang2025shape, yu2026recondrive}, optical flow \citep{lin2025movies, yang2026neoverse, he2026dynamicvggt}, or pre-trained trackers  \citep{fei2024driv3r,chen2025dggt}, remains a highly challenging task. Recent works have explored fully self-supervised, scene-optimized strategies to decompose dynamic objects. Specifically, methods like S3Gaussian \cite{huang2024textit} and EvoGS \cite{asiimwe20254d} capture scene dynamics by learning a spatial-temporal hexplane representation. Despite their high fidelity, these methods require time-consuming per-scene optimization, limiting their scalability. To address this, another group of approaches turns to exploring the feed-forward unsupervised dynamic reconstruction. Some methods, such as V-DPM \cite{sucar2026v} and BTimer \cite{liang2024feed}, implicitly learn dynamic reconstruction at given target timestamps utilizing time-conditioned transformer blocks. In contrast, other feed-forward methods, e.g., Flux4D \cite{wang2025flux4d}, 4DGT \cite{xu20254dgt}, and STORM \cite{yang2024storm}, explicitly predict motion parameters for each 3D Gaussian. However, these approaches only estimate instantaneous motion vectors at the current timestamp and move Gaussians with a simplified linear motion assumption, which struggles with accumulated trajectory drift over extended temporal horizons.

\textbf{Future Scene Prediction}
Future scene prediction has been extensively explored in video generation models \citep{ho2022video, hong2022cogvideo, blattmann2023stable, tian2024visual, gao2024vista, wu2025video}. Representative foundation models, such as Sora \citep{liu2024sora}, CogVideoX \citep{yang2024cogvideox}, Cosmos \citep{ali2025world}, and Wan \citep{wan2025wan}, have demonstrated remarkable visual synthesis capabilities.
To enhance view consistency and structural stability, recent works like Gen3R\citep{huang2026gen3r}, FantasyWorld \citep{dai2025fantasyworld}, and Geometry Forcing \citep{wu2025geometry} align diffusion models with geometry-aware structures from 3D reconstruction. DINO-World \citep{baldassarre2025back}, DINO-Foresight\citep{karypidis2026dino}, and VGGT-World \citep{sun2026vggt} directly use frozen geometry-foundation features as the latent state and model their future evolution. Unlike previous approaches that rely on implicit latent evolution, our method explicitly extrapolates future states by jointly predicting future camera poses and dynamic motions with better controllability.




\begin{figure}[tp] 
  \centering
  \includegraphics[width=1\columnwidth]{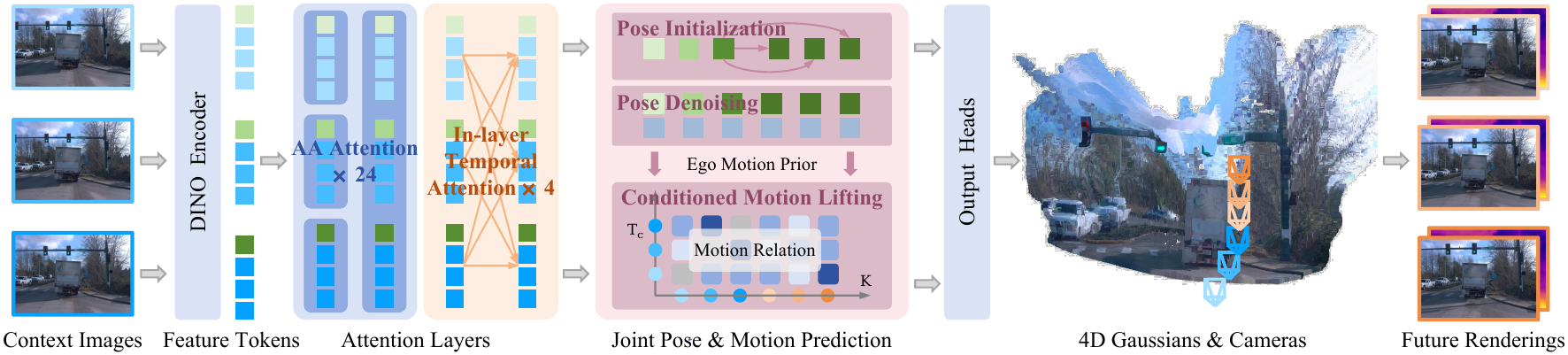} 
  \caption{\textbf{Framework of Envision4D.} Given a sequence of context images, Envision4D predicts 4D Gaussians and all target camera poses. The motion awareness of feature tokens is first enhanced via In-layer Temporal Attention. Subsequently, Joint Pose-Motion Prediction is applied to enable future pose estimation through iterative denoising, alongside non-linear motion generation via conditioned motion lifting. The generated tokens are then decoded to render novel future views.}
  \vspace{-10pt}
  \label{pipline} 
\end{figure}

\section{Method}

The methodology is organized as follows. We first present the task formulation (Sec. \ref{sec3.1}). Subsequently, we introduce Envision4D, a novel framework designed for effective self-supervised 4D Gaussian Splatting. Envision4D consists of three key components: a frozen VGGT encoder with in-layer temporal attention to formulate a motion-biased feature space (Sec. \ref{sec3.2}), a joint pose-motion prediction module to facilitate future scene extrapolation in a pose-free manner (Sec. \ref{sec3.3}), and a progressive training strategy to optimize the unsupervised motion learning process (Sec. \ref{sec3.4}).

\subsection{Task Formulation}
\label{sec3.1}
As shown in Fig. \ref{pipline}, we formulate future scene reconstruction as a pose-free and self-supervised motion learning task. Given a sequence of $T_c$ context images, the proposed Envision4D aims to reconstruct the dynamic scene and predict its evolution over a future horizon of $T_f$ frames. For each observed frame $i \in [1, T_c]$, the model predicts a depth map $D_i$ and a set of Gaussian attributes $G_i$. Simultaneously, for all timestamps $j \in [1, T_c + T_f]$, the model estimates the camera parameters $P_j$ and the time-conditioned velocity $V_{i,j}$ from source frame $i$ to target frame $j$ (where $j \neq i$).


\noindent\textbf{Future Pose-free Extrapolation.} To predict the scene state at any target timestamp $j \in [1, T_c + T_f]$, we warp the Gaussian centers from a source timestamp $i$ using a time-conditioned velocity, rather than constant linear motion across different target frames:
\begin{equation}\mu_{i \to j} = \mu_i + \mathbf{v}_{i,j} \cdot (j-i), \quad i \in [1, T_c], j \in [1,  T_c + T_f]\label{eq_mu}\end{equation}
where $\mu \in \mathbb{R}^3$ denotes the 3D position, $j-i$ acts as a scaling factor to normalize the velocity learning across varying time gaps. By rendering the aggregated Gaussians via the estimated poses $P_j$ from Sec. \ref{sec3.3}, Envision4D enables high-quality future view synthesis in a truly pose-free manner.

\subsection{Motion Awareness via In-Layer Temporal Attention}
\label{sec3.2}
Our model is built upon the VGGT backbone \cite{wang2025vggt}, which processes a sequence of images $\mathbf{I} \in \mathbb{R}^{T_c \times 3 \times H \times W}$ into frame-wise tokens via DINOv2 \cite{oquab2023dinov2} and 24 layers of Alternating-Attention (AA). While concurrent works \cite{he2026dynamicvggt, yang2026neoverse} typically append auxiliary motion modules on the top of the output frozen tokens, such post-refining paradigms limit motion learning to the newly added shallow blocks, failing to exploit the deep hierarchical priors of the encoder.

In contrast, we propose an In-Layer Temporal Attention. By embedding efficient temporal attention blocks directly into the intermediate stages of the frozen VGGT encoder, we empower the subsequent frozen AA layers to progressively propagate and reinforce the learned motion cues throughout the feature extraction process. Let $\mathbf{F}_l$ be the output of the $l$-th AA layer. Formally, we have:
\begin{equation}
 \mathbf{F}_{l+1} = \mathrm{AA}_{l+1}(\mathrm{TAttn}(\mathbf{F}_l)), \quad l \in \{4, 11, 17, 23\}
\end{equation}
where $\mathrm{TAttn}$ performs attention across the temporal dimension. The tokens from global attention and our temporal attention are then concatenated as motion tokens $\mathbf{F}^M$ for subsequent processing.

However, injecting newly initialized modules into the intermediate stages of a frozen network inherently alters the feature distribution for subsequent layers. This risks severe degradation of the original feature space, which fundamentally explains why existing approaches settle for sub-optimal post-layer refinements. To resolve this dilemma and stabilize the intermediate representations, we introduce a self-distillation supervision strategy (detailed in the training loss part of Sec. \ref{sec3.4}).

\subsection{Future Extrapolation via Joint Pose-Motion Prediction}
\label{sec3.3}
\textbf{Future Pose Prediction.} 
The first step for pose-free future extrapolation is to generate the subsequent camera poses. Previous video generation methods \cite{zhang2025epona} typically utilize Diffusion Transformers \citep{peebles2023scalable} to forecast future $T_f$-frame pose trajectories, which often incur substantial GPU memory overhead and are notoriously difficult to regress. Instead, we formulate the future pose prediction directly within a compact geometric feature space.

Given the aggregated camera pose tokens across all context frames, let $\mathbf{z}_{T_c} \in \mathbb{R}^{C}$ denote the camera token of the last observed frame. We initialize the predictions for the subsequent $T_f$ unknown frames by adding a learnable offset $\boldsymbol{\delta} \in \mathbb{R}^{C}$ to $\mathbf{z}_{T_c}$:
\begin{equation}
\mathbf{z}_j = \mathbf{z}_{T_c} + \boldsymbol{\delta}, \quad j \in (T_c, T_c + T_f]
\end{equation}
where $\boldsymbol{\delta}$ serves as an initial noisy seed for the extrapolated poses.

The known pose tokens, together with the initialized future tokens, form the full sequence representation $\mathbf{F}^{\text{cam}} \in \mathbb{R}^{(T_c+T_f) \times C}$. Additionally, to endow each token with positional awareness along the temporal axis, we compute a 1D sinusoidal time embedding for all frame index:
\begin{equation}
\mathbf{e}_i = \mathrm{Linear}(\mathrm{SinEmb}(i)) \in \mathbb{R}^{C}, \quad i \in \{1, \ldots, T_c+T_f\}
\end{equation}
The sequence tokens are then concatenated with corresponding time embeddings and passed through a stack of self-attention blocks. This mechanism allows the model to iteratively refine the noisy future predictions by globally conditioning on the observed history and their time embeddings:
\begin{equation}
\mathbf{Z}^{(l+1)} = \mathrm{SelfAttn}(\mathbf{Z}^{(l)}), \quad \text{where} \ \mathbf{Z}^{(0)} = [\mathbf{F}^{\text{cam}}, \mathbf{E}]
\label{eq_selfattn}
\end{equation}
where $\mathbf{E} = [\mathbf{e}_1, \dots, \mathbf{e}_{T_c+T_f}] \in \mathbb{R}^{(T_c+T_f) \times C}$ represents the time embedding matrix, and $\mathbf{Z}^{(l)}$ denotes the updated token sequence at the $l$-th attention layer.
After that, the refined camera tokens are fed into a pre-trained, frozen camera head to decode the extrapolated camera parameters $P$.

\textbf{Conditioned Motion Lifting.}
Unlike bidirectional motion in STORM \cite{yang2024storm}, we directly model time-conditioned velocities by formulating motion prediction as a dynamic relational mapping between current observations and other target states. This design enables the network to adaptively capture complex dynamics, avoiding the error accumulation in conventional linear extrapolation.

We explicitly lift current motion observations into a temporal grid by correlating each context token with $K$ alternative timestamps, where $K = T_c + T_f - 1$ denotes all sequence time steps excluding the current reference frame. Unlike naive regression, the contextual features are modulated to represent a motion relational mapping. Specifically, given the learned motion tokens $\mathbf{F}^M \in \mathbb{R}^{T_c \times N \times C}$, let $f_{i,n} \in \mathbb{R}^C$ denote a specific token at context frame $i$ and spatial location $n$. Leveraging the motion priors $\mathbf{E} \in \mathbb{R}^{K \times C}$, which distill both temporal dependencies and ego-motion from the preceding \textit{Future Pose Prediction} stage, the future target motion feature $m_{i,j,n} \in \mathbb{R}^C$ is defined as the context vector $f_{i,n}$ scaled by the corresponding motion prior $\mathbf{E}_j$:
\begin{equation}
m_{i,j,n} = \mathbf{E}_j \odot f_{i,n}
\end{equation}
This feature modulation ensures the resulting velocity field is physically consistent with future timestamps and ego-motion, leading to more robust future scene extrapolation.

Finally, we flatten the temporal dimensions of the resulting volume $\mathbf{M} \in \mathbb{R}^{T_c \times K \times N \times C}$ and pass it through a DPT head to yield the final velocity predictions $V \in \mathbb{R}^{(T_c \times K) \times N \times 3}$.

\subsection{Progressive Training Strategy}
\label{sec3.4}
Future extrapolation inherently suffers from much more severe error accumulation than interpolation, with errors growing sharply as the number of extrapolated frames increases. This makes unsupervised velocity learning highly susceptible to model collapse. To stabilize the learning process, we introduce a progressive training strategy that regularizes the network across both the rendering mechanisms and the extrapolation optimization length.

\textbf{Geometric Warm-up and Self-Exclusive Motion Learning.} In the early training stages, we mainly focus on static geometry and pose optimization by enforcing only reconstruction on the context frames. Once a stable geometric foundation is established, we facilitate unsupervised motion learning using a self-exclusive rendering strategy. Specifically, during the rendering phase, we remove the target frame's own Gaussians. This strategy is crucial to prevent model degeneration, where the network might otherwise bypass actual motion learning by exploiting a trivial identity mapping of the target frame's appearance. By amplifying motion-induced misalignments, it forces the network to strictly distinguish between static and dynamic elements, ensuring robust velocity estimation.

\textbf{Progressive Extrapolation Weighting.} 
To further stabilize highly uncertain extrapolation process, we employ an exponentially decaying weighting scheme that down-weights future frames relative to context frames. For a frame at time index $t$, its unnormalized loss weight is computed as:
\begin{equation}w_t = \exp(-\alpha \cdot \max(0, t - T_c + 1))\end{equation}
where the decay rate $\alpha$ linearly anneals from an initial value $\alpha_0$ to $0$ over $S_{max}$ training steps. Crucially, $w_t$ is dynamically normalized to maintain a mean of 1 across the sequence, ensuring a stable gradient scale throughout the annealing process.

\textbf{Training Loss.} The primary training signal is driven by the image rendering loss, where we combine Mean Squared Error (MSE) and Learned Perceptual Image Patch Similarity (LPIPS) losses:
\begin{equation}
    \mathcal{L}_\mathrm{rgb} = \mathcal{L}_\mathrm{MSE} + \lambda_\mathrm{lpips} \mathcal{L}_\mathrm{LPIPS}
\end{equation}
Furthermore, we propose a self-distillation loss to regularize scene geometry. It functions as a soft regularizer that provides future geometric supervision, and crucially, aligns the updated feature space with the pre-trained one to prevent feature degradation. Specifically, we directly leverage the outputs of a frozen VGGT backbone as pseudo-supervision, formulated as:
\begin{equation}
    \mathcal{L}_\mathrm{cam} = \| P - P_\mathrm{vggt} \|_1, \quad \mathcal{L}_\mathrm{depth} = \| D - D_\mathrm{vggt} \|_1
\end{equation}
where $P_\mathrm{vggt}$ and $D_\mathrm{vggt}$ denote the camera matrices and depth maps predicted by the VGGT network.


\vspace{-2pt}
\section{Experimental Results}
\label{sec:result}
\vspace{-2pt}

\begin{table}[tp]
\centering
\caption{\textbf{Comparison to state-of-the-art methods on the Waymo dataset.} We evaluate our method against existing approaches, with the two most relevant baselines re-implemented under identical settings for a fair comparison. Inference speed is measured on a single A100 GPU.}
\label{tab:main_results}
\resizebox{1\columnwidth}{!}{
\begin{tabular}{lcccccc} 
\toprule
\multirow{2}{*}{Methods} & \multicolumn{3}{c}{Render Quality} & \multicolumn{1}{c}{Inference Speed} & \multicolumn{2}{c}{Capability} \\ 
\cmidrule(lr){2-4} \cmidrule(lr){5-5} \cmidrule(lr){6-7} 
& PSNR $\uparrow$ & SSIM $\uparrow$ & D-RMSE $\downarrow$ & Time $\downarrow$ & Pose-free & Unsup. Dynamic \\
\midrule
\multicolumn{7}{l}{\textit{Per-scene Optimization Methods}} \\ 
PVG \citep{chen2026periodic} & 22.38 & 0.661 & 13.01 & 27 min & $\times$ & \checkmark \\
DeformableGS \citep{yang2024deformable} & 25.29 & 0.761 & 14.79 & 29 min & $\times$ & \checkmark \\
\midrule
\multicolumn{7}{l}{\textit{Generalizable Feed-forward Methods}} \\ 
DepthSplat \citep{xu2025depthsplat} & 23.26 & 0.696 & 10.05 & 0.11 s & $\times$ & $\times$ \\
NoPoSplat \citep{ye2024no} & 24.31 & 0.751 & 9.08 & 23.22 s & \checkmark & $\times$ \\
STORM \citep{yang2024storm} & 26.38 & 0.794 & 5.48 & 0.18 s & $\times$ & \checkmark \\
DGGT \citep{chen2025dggt} & 27.41 & 0.846 & 3.47 & 0.39 s & \checkmark & $\times$ \\
\midrule
\multicolumn{7}{l}{\textit{Our Reproductions \& Method}} \\ 
STORM* \citep{yang2024storm} & 26.19 & 0.798 & 6.13 & \textbf{0.12 s} & $\times$ & \checkmark \\
DGGT* \citep{chen2025dggt} & 24.38 & 0.756 & 7.67 & 0.56 s & \checkmark & $\times$ \\
\rowcolor{pink!20}
Ours & \textbf{27.81} & \textbf{0.816} & \textbf{3.98} & 0.37 s & \checkmark & \checkmark \\
\bottomrule
\end{tabular}}
\end{table}

\begin{figure}[tp] 
  \centering
  \vspace{-8pt}
  \includegraphics[width=1\columnwidth]{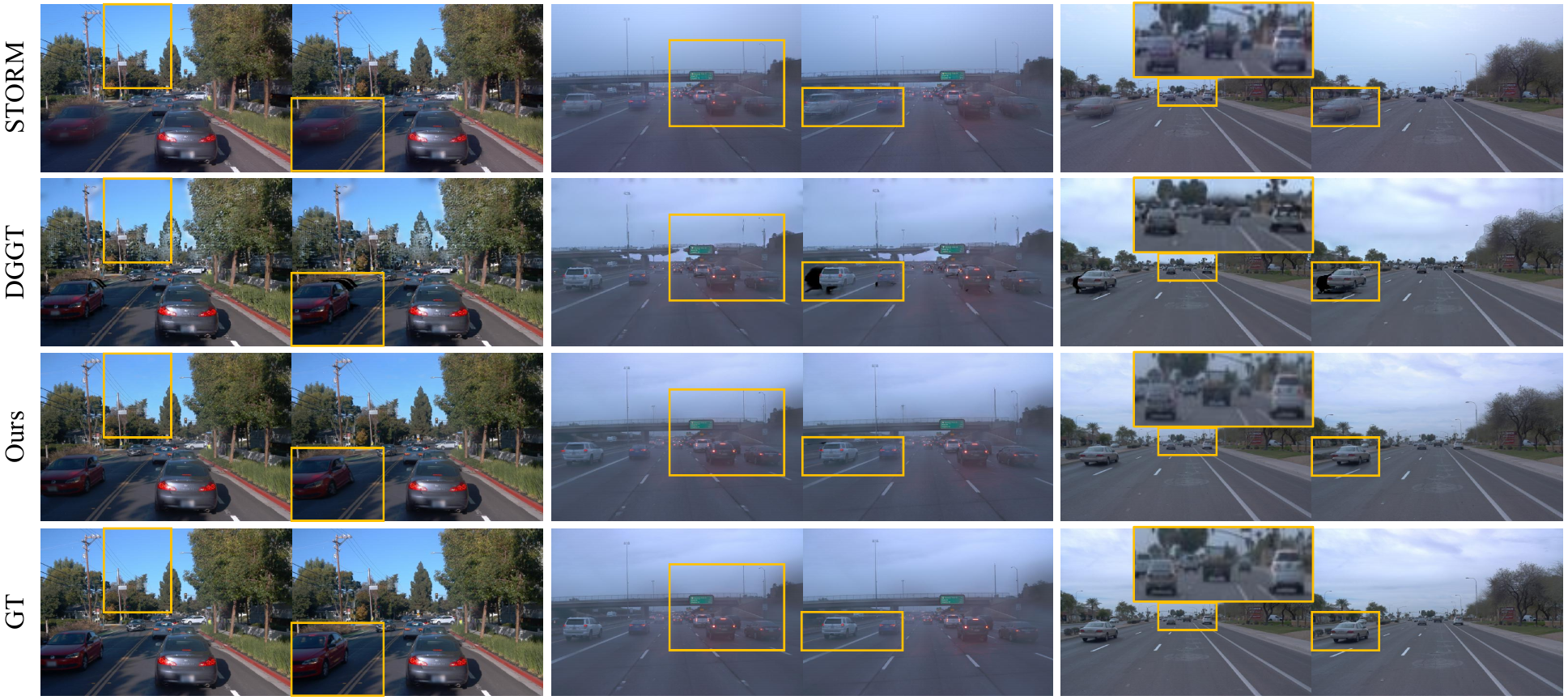} 
  \caption{\textbf{Qualitative comparison on Waymo dataset.} Even without explicit motion guidance and known future poses, our model can well handle large movements under various conditions.}
  \vspace{-9pt}
  \label{fig:com1} 
\end{figure}

\vspace{-2pt}
\subsection{Experimental Setup}
\vspace{-2pt}

\textbf{Setup.} 
We evaluate our method on the official validation splits of Waymo \cite{sun2020scalability} and nuScenes \cite{caesar2020nuscenes} datasets. For each validation clip, we condition on $T_c$ frames to generate a full sequence of $T_c+T_f$ frames (where $T_c=T_f=2$ by default), challenging the model to deduce future motions from sparse observations. 
\textit{During comparative evaluation, we assess exclusively the extrapolated future frames. For ablation studies, we evaluate the full sequence, including \textbf{context frames rendered via solely cross-frame Gaussians}, to validate overall motion alignment.}

\vspace{-1.5pt}
\textbf{Implementation Details.} 
Our framework utilizes GSplat \cite{Ye2024gsplatAO} as the highly efficient Gaussian Splatting rendering backend. All input video frames are resized to a resolution of $350 \times 518$. The model is trained on a single A100 GPU with batch size 2. The total training process spans 100K iterations.

\vspace{-2pt}
\subsection{Comparison Study}
\vspace{-2pt}

\textbf{Rendering Quality.} 
In Tab. \ref{tab:main_results}, we compare with both per-scene optimization and generalizable feed-forward methods. Specifically, we re-implement the two most relevant baselines, STORM and DGGT, to our extrapolation setting. As presented, the performance of DGGT drops sharply during future extrapolation. This degradation occurs because DGGT relies on a frozen off-the-shelf tracker to calculate motions and interpolates new camera poses between observed frames, extending this mechanism to extrapolate introduces severe black shadows and inaccurate pose shifts. While STORM maintains comparable performance, our Envision4D outperforms baselines in rendering quality under the challenging constraints of future pose-free and unsupervised dynamic learning. Fig. \ref{fig:com1} shows qualitative comparisons, where STORM struggles with ghosting artifacts, and DGGT yields trajectory deviations under extrapolation. In contrast, Envision4D accurately captures large displacements without trailing artifacts, delivering a much sharper overall appearance. 
We also achieve competitive results on nuScenes (Tab. \ref{tab:nu_results}). Crucially, Envision4D is evaluated under the challenging extrapolation setting, while baselines report their original, interpolation-primary results.

\begin{table}[tb]
    \centering
    \begin{minipage}[t]{0.45\textwidth} 
        \centering
        \caption{\textbf{Comparison to state-of-the-art methods on the nuScenes dataset.} The results are cited from original papers.}
        \label{tab:nu_results}
        \resizebox{\linewidth}{!}{
        \begin{tabular}{lccc}
            \toprule
            Method & PSNR $\uparrow$ & SSIM $\uparrow$ & LPIPS $\downarrow$ \\
            \midrule
            STORM \citep{yang2024storm} & 24.54 & 0.784  &0.267 \\
            DGGT \citep{chen2025dggt} & 26.63 & 0.813 & \textbf{0.122} \\
            Ours & \textbf{26.86}& \textbf{0.815}& 0.164\\
            \bottomrule
        \end{tabular}}
    \end{minipage}
    \hfill 
    \begin{minipage}[t]{0.50\textwidth} 
        \centering
        \caption{\textbf{Camera pose estimation on the Waymo and nuScenes datasets.} Metrics are evaluated on the full sequence.}
        \label{tab:auc_comparison}
        \resizebox{\linewidth}{!}{
        \begin{tabular}{lccc}
            \toprule
            \multirow{2}{*}{Method} & Future & Waymo & nuScenes \\
            & camera & AUC@30 $\uparrow$ & AUC@30 $\uparrow$ \\
            \midrule
            VGGT \citep{wang2025vggt} & $\times$ & 78.58 & 76.99 \\
            Ours & \checkmark & \textbf{79.49} & \textbf{78.03} \\
            \bottomrule
        \end{tabular}}
    \end{minipage}
\end{table}

\begin{table}[tp]
    \centering
    \vspace{-13pt}
    \caption{\textbf{Quantitative comparison under varying context ($T_c$) and future ($T_f$) frames.} Notably, our method at $T_f=6$ even achieves competitive performance compared to STORM at $T_f=2$.}
    \label{tab:fut} 
    \resizebox{0.76\linewidth}{!}{
    \begin{tabular}{cccccccc} 
        \toprule
        \multirow{2}{*}{$T_c$} & \multirow{2}{*}{$T_f$} & \multicolumn{2}{c}{PSNR $\uparrow$} & \multicolumn{2}{c}{SSIM $\uparrow$} & \multicolumn{2}{c}{LPIPS $\downarrow$} \\
        \cmidrule(lr){3-4} \cmidrule(lr){5-6} \cmidrule(lr){7-8}
        & & STORM \citep{yang2024storm} & Ours & STORM \citep{yang2024storm} & Ours & STORM \citep{yang2024storm} & Ours \\
        \midrule
        2 & 2 & 26.19 & \textbf{27.81} & 0.798 & \textbf{0.816} & 0.242 & \textbf{0.159} \\
        2 & 4 & 25.55 & \textbf{26.87} & 0.773 & \textbf{0.790} & 0.263 & \textbf{0.170} \\
        2 & 6 & 24.41 & \textbf{26.21} & 0.753 & \textbf{0.771} & 0.346 & \textbf{0.192} \\
        3 & 6 & 24.62 & \textbf{26.46} & 0.752 & \textbf{0.782} & 0.302 & \textbf{0.187} \\
        \bottomrule
    \end{tabular}}
  \vspace{-9pt}
\end{table}

\begin{figure}[ht] 
  \centering
  \vspace{-6pt}
  \includegraphics[width=1\columnwidth]{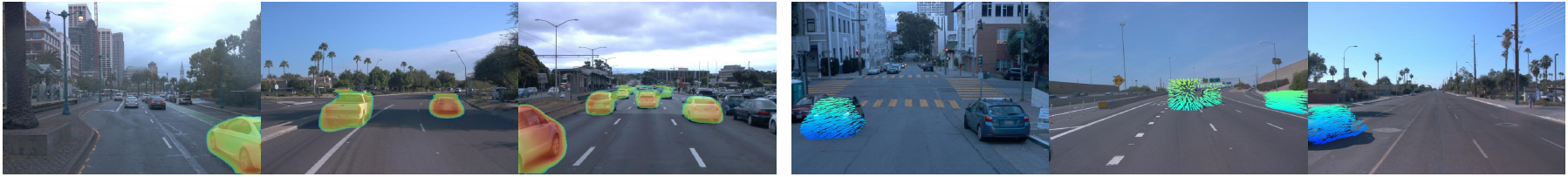} 
  \caption{\textbf{Visualization of dynamic masks and predicted velocities.} Left: dynamic mask. Right: scene flow. We overlay the original image on each example to enhance clarity.}
  \vspace{-6pt}
  \label{fig:motion_flow} 
\end{figure}

\textbf{Camera Pose Estimation.}
We evaluate our method for camera pose estimation on the two datasets. In particular, VGGT \citep{wang2025vggt} receives all target images as input, whereas our model requires only two frames to predict both current and future camera poses. As shown in Tab. \ref{tab:auc_comparison}, our method attains higher AUC@30 accuracy despite the added challenge of forecasting future trajectories. This performance gain over our VGGT-derived pseudo-labels stems from the integration of image reconstruction loss, which serves a joint optimization of scene appearance and camera trajectories. Consequently, our framework can also function as an unsupervised refinement step to boost VGGT accuracy in the absence of camera calibrations.

\textbf{Context and Extrapolation Lengths.}
Unlike interpolation where motion errors are typically constrained between observations, extrapolation is highly ill-posed, with errors accumulating sharply as the number of extrapolated future frames increases. 
As analyzed in Tab. \ref{tab:fut}, our method yields stable high-fidelity rendering across extended extrapolation horizons. Notably, our long-term prediction ($T_f=6$) yields even higher PSNR and LPIPS quality than STORM's short-term output ($T_f=2$), demonstrating exceptional robustness against temporal error accumulation. Additionally, extending context frames further enhances extrapolation capability by providing richer dynamic cues.

\vspace{-1.5pt}
\subsection{Ablation Study}
\vspace{-1.5pt}

We analyze the contribution of each proposed component in Tab. \ref{tab:ablation}. A \textit{Future Pose (FP) Prediction} module, coupled with linear bidirectional velocity estimation, serves as our baseline. Building upon these, the integration of \textit{Conditioned Motion Lifting (CML)} yields a substantial improvement, highlighting the necessity of our time-conditioned velocity for future extrapolation. Furthermore, our proposed \textit{In-layer TAttn} outperforms the conventional \textit{Post-layer TAttn}, which implies that integrating temporal attention deeply within the network layers facilitates better motion-aware feature fusion than late-stage processing. Finally, training the model directly without our \textit{Progressive Training Strategy (Prog. Train)} causes a significant performance drop, underscoring its crucial role in stabilizing the unsupervised motion learning in scene extrapolation.

\vspace{-2.5pt}
\subsection{Qualitative Results and Applications}
\vspace{-2.5pt}

\textbf{Dynamic Segmentation and Flow Estimation.} 
As visualized in Fig. \ref{fig:motion_flow}, our model accurately distinguishes dynamic elements from static backgrounds and generates high-fidelity 3D velocities. Despite the absence of explicit motion guidance, Envision4D demonstrates the capability to differentiate between a moving vehicle and a nearby stationary one, producing accurate dynamic masks and motion flows that are valuable for downstream autonomous driving tasks.

\begin{figure}[tbp] 
  \centering
  \vspace{-6pt}
  \includegraphics[width=1\columnwidth]{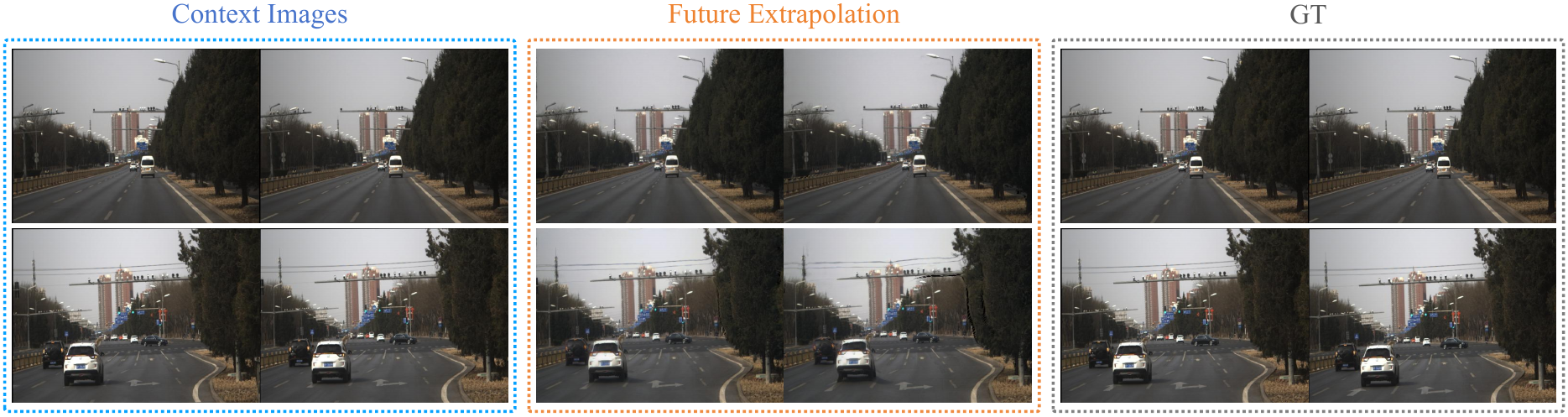} 
  \caption{\textbf{Qualitative results on in-the-wild data.} Envision4D is capable to generate reliable future extrapolation directly from uncalibrated open-world context images.}
  \vspace{-13pt}
  \label{fig:wild} 
\end{figure}

\textbf{In-the-wild Scene Reconstruction.} 
To demonstrate the generalization capability, we extend evaluations to challenging in-the-wild scenarios with entirely unknown camera parameters. As shown in Fig. \ref{fig:wild}, despite the high uncertainty of unconstrained camera and object movements, Envision4D achieves robust future forecasting with superior rendering fidelity and geometric alignment.


\begin{table}[tbp]
    \centering
    \caption{\textbf{Ablation study on Waymo dataset.} \textit{Post-layer TAttn} denotes appending the temporal attention module after frozen backbone, whereas our proposed \textit{In-layer TAttn} integrates it inside.}
    \label{tab:ablation}
    \resizebox{0.88\linewidth}{!}{
    \begin{tabular}{ccccc|ccc}
        \toprule
        \textit{FP} & \textit{CML} & \textit{In-layer TAttn} & \textit{Post-layer TAttn} & \textit{Prog. Train} & PSNR $\uparrow$ & SSIM $\uparrow$ & LPIPS $\downarrow$ \\
        \midrule
        \checkmark &            &            &            & \checkmark & 25.41 & 0.796 & 0.198 \\
        \checkmark & \checkmark &            &            & \checkmark & 27.29 & 0.812 & 0.167 \\
        \checkmark & \checkmark &            & \checkmark & \checkmark & 28.01 & 0.824 & 0.158 \\
        \checkmark & \checkmark & \checkmark &            &            & 27.89 & 0.829 & 0.160 \\
        \checkmark & \checkmark & \checkmark &            & \checkmark & \textbf{28.83} & \textbf{0.849} & \textbf{0.145} \\
        \bottomrule
    \end{tabular}}
\end{table}

\vspace{-3pt}
\section{Limitations and Conclusion}
\label{sec:conclusion}
\vspace{-3pt}

\textbf{Limitations.} 
While our work exhibits stronger self-supervised motion learning quality and relieves the model from inflexible extrapolation constraints, it presents certain limitations. First, motion estimation for faraway, rapidly approaching objects is constrained by the extreme sparsity of input visual cues. Second, inherent to our reconstruction-based nature, the model struggles to hallucinate entirely unseen regions, meaning it cannot forecast as far into the future as generative models. In future work, we aim to incorporate generative priors to enable longer-horizon extrapolation.

\textbf{Conclusion.} In this paper, we introduce Envision4D, a novel 4DGS model for future scene extrapolation without explicit motion guidance or predefined future cameras. We propose \textit{Joint Pose-Motion Prediction} to infer all target camera poses and corresponding non-linear Gaussian velocities, utilizing \textit{In-layer Temporal Attention} to strengthen temporal perception and facilitate motion learning. Additionally, a \textit{Progressive Training Strategy} stabilizes the unsupervised learning process and mitigates error accumulation. Extensive experiments demonstrate state-of-the-art performance in dynamic scene reconstruction and strong generalization capability on in-the-wild data.


\clearpage


\bibliography{example}  

\clearpage

\appendix
\counterwithin{table}{section}
\counterwithin{figure}{section}
\input{appendix.tex}

\end{document}

%% file: appendix.tex
\section{Implementation Details}

\textbf{Model Architecture.}
In our model, each Gaussian primitive is parameterized as $G = \{ \mu, r, s, c, \alpha \}$, where $\mu \in \mathbb{R}^3$ denotes the 3D center position, $r \in \mathbb{R}^4$ is the rotation quaternion, $s \in \mathbb{R}^3$ represents the scaling factor, $c \in \mathbb{R}^3$ is the color, and $\alpha \in \mathbb{R}$ signifies the opacity. 
Specifically, the 3D positions $\mu$ are obtained by back-projecting the predicted depth into the 3D space using the estimated camera metrics. 
The color attributes $c$ are normalized into $[0, 1]$ via a sigmoid activation function. 

For Eq.~\ref{eq_selfattn} specified in the \textit{Future Pose Prediction} module, the self-attention block comprises $2$ layers with $16$ attention heads. 
To optimize the predicted poses, this block is executed recurrently for $4$ sequential refinement passes. 
In each pass, the shared self-attention block updates the token sequence $\mathbf{Z}^{(l)}$, allowing the model to iteratively infer smooth and dynamically consistent ego motions.

\textbf{Training and Optimization.}
We train our framework using the Adam optimizer coupled with a cosine learning rate scheduler, setting the initial learning rate to $1\times 10^{-4}$. 
The balancing hyperparameters for the loss functions are empirically set to $\lambda_\mathrm{LPIPS} = 0.05$ and $\lambda_\mathrm{cam} = 5.0$. 
For our \textit{Progressive Training Strategy}, the geometric warm-up stage spans the first $2,500$ iterations. 
Subsequently, the maximum step $S_{\max}$ in the progressive extrapolation weighting phase is set over $50\text{K}$ iterations, with the initial decay rate $\alpha_0$ initialized to $1.0$.

\section{Additional Results}
\textbf{Cross-frame Scene Reconstruction.}
Different from previous works that evaluate reconstruction performance using the input frames themselves, which may easily lead to overfitting static geometry rather than assessing motion, we introduce a more strict cross-frame rendering evaluation to thoroughly examine the model's capability in estimating velocities between context frames. To ensure fairness, the baseline model is also enhanced with our self-exclusive motion learning. 
Meanwhile, since the backward rendering inevitably introduces invisible regions near the image boundaries, we apply a boundary cropping during evaluation. 
As shown in Tab.~\ref{tab:cross_frame_new}, our model significantly outperforms baseline method in both forward and backward rendering, firmly demonstrating its superior capability to extract accurate motion from given context clues.

\begin{table}[htp]
    \centering
    \vspace{-7pt}
    \caption{\textbf{Cross-frame scene reconstruction performance.} The notations $1\rightarrow0$ and $0\rightarrow1$ denote the backward (warping frame 1 to render frame 0) and forward (warping frame 0 to render frame 1) synthesis, respectively.}
    \label{tab:cross_frame_new} 
    \resizebox{0.8\linewidth}{!}{
    \begin{tabular}{lcccccc} 
        \toprule
        \multirow{2}{*}{Method} & \multicolumn{3}{c}{1$\rightarrow$0} & \multicolumn{3}{c}{0$\rightarrow$1} \\
        \cmidrule(lr){2-4} \cmidrule(lr){5-7}
        & PSNR $\uparrow$ & SSIM $\uparrow$ & LPIPS $\downarrow$ & PSNR $\uparrow$ & SSIM $\uparrow$ & LPIPS $\downarrow$ \\
        \midrule
        STORM \citep{yang2024storm} & 27.62& 0.836& 0.237& 27.97&0.838 &0.218 \\
        Ours                        &30.36 &0.886 &0.122&30.71 & 0.891&0.113 \\
        \bottomrule
    \end{tabular}}
    \vspace{-2pt}
\end{table}

\begin{figure}[tp] 
  \centering
  \includegraphics[width=1\columnwidth]{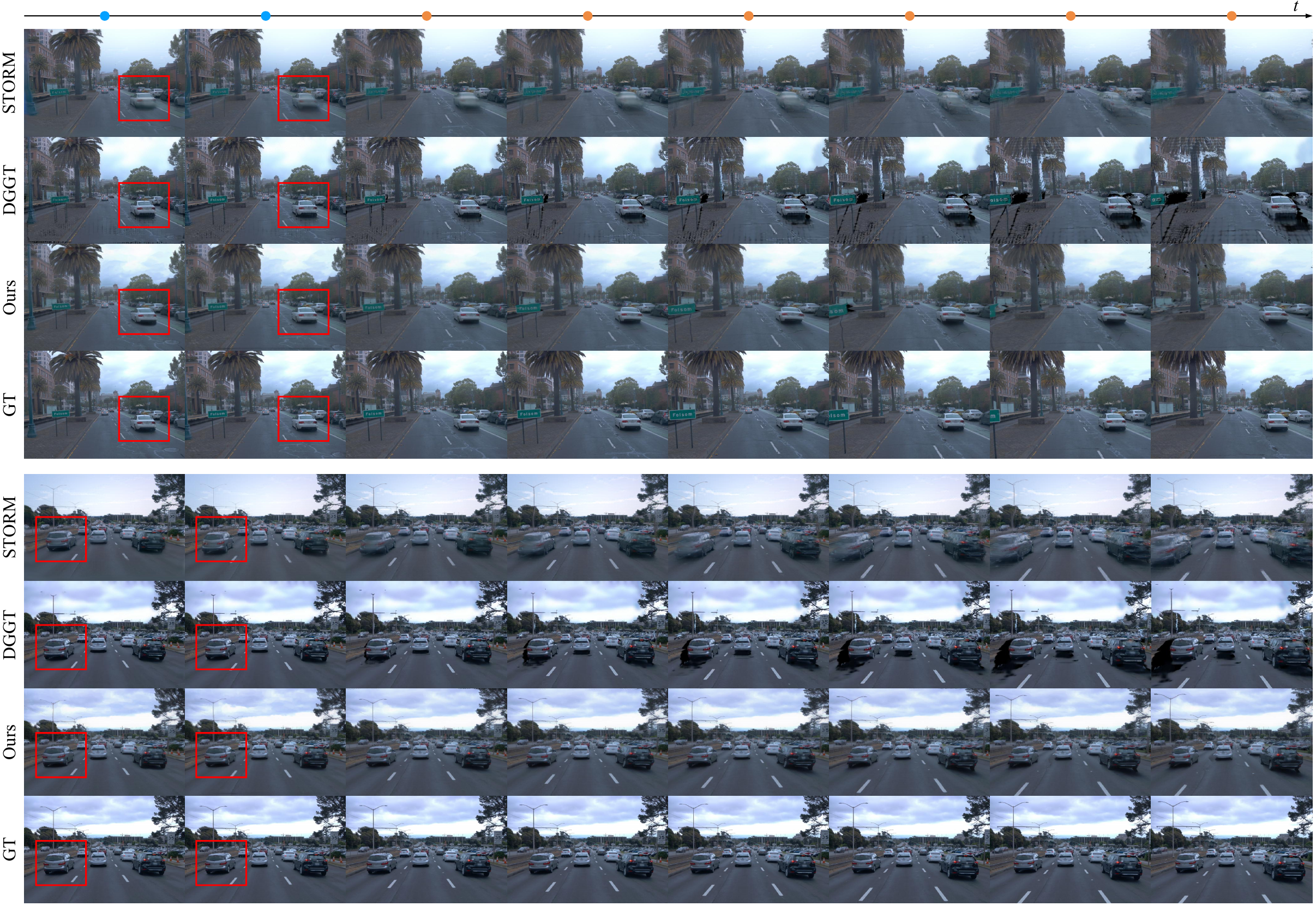} 
  \caption{\textbf{Qualitative comparisons of long-term future extrapolation.} The blue and orange dots denote the input observation frames and extrapolated future frames, respectively. The challenging dynamic objects are highlighted in red boxes across the input frames. It is noted that our model achieves significantly more stable and temporally consistent future extrapolation, even without extra motion guidance and ground-truth future poses.}
  \label{fig:ex2-6} 
\end{figure}

\textbf{Long-term Extrapolation.}
To further analyze the error accumulation inherent in future extrapolation, we provide qualitative comparisons across $8$ continuous frames in two distinct scenes. In each sequence, the first $2$ frames serve as input observations, while the subsequent $6$ frames represent the extrapolated future. As visualized in Fig. \ref{fig:ex2-6}, the motion drift of dynamic objects becomes progressively severe as the temporal extrapolation distance increases. Specifically, under an unconstrained extrapolation setting, the motion estimation of the unsupervised approach, i.e., STORM~\citep{yang2024storm}, becomes highly unstable. Also, restricted by its linear velocity assumption, STORM produces severe ghosting artifacts for dynamic objects at distant frames. On the other hand, although DGGT~\citep{chen2025dggt} utilizes a pre-trained tracker to capture object motion, it still inevitably suffers from trajectory deviation, as seen with the right black vehicle in the second scene. Furthermore, during future extrapolation, DGGT can only derive future novel-view poses through naive linear pose extrapolation. This oversimplified assumption fails to capture complex real-world camera trajectories, resulting in significant ego-pose drift (as illustrated in the second row of Fig. \ref{fig:ex2-6}) and a sharp decline in overall extrapolation accuracy. In contrast, our method achieves notably stable and temporally consistent motion estimation, successfully maintaining high-fidelity reconstruction even at distant future.

\textbf{Ablation Study.}
We further study the effects of different components in the \textit{Progressive Training Strategy}. As presented in Tab. \ref{tab:ablation_prog}, the geometric warm-up stage consistently improves performance, since enhancing static texture information at the beginning of training strengthens textual details and establishes a solid geometric foundation. Additionally, progressively increasing the loss weights of extrapolated frames yields substantial gains, proving its efficacy in stabilizing unsupervised motion learning. When both strategies are combined, the model achieves the best performance across all metrics, showcasing their complementary nature in improving dynamic scene extrapolation.

\begin{table}[tp]
    \centering
    \vspace{-3pt}
    \caption{\textbf{Ablation study on Progressive Training Strategy.} \textit{Warm-up} denotes the geometric warm-up stage and \textit{Prog. Weighting} represents the progressive extrapolation weighting strategy.}
    \label{tab:ablation_prog}
    \resizebox{0.65\linewidth}{!}{ 
    \begin{tabular}{cc|ccc}
        \toprule
        \textit{Warm-up} & \textit{Prog. Weighting} & PSNR $\uparrow$ & SSIM $\uparrow$ & LPIPS $\downarrow$ \\
        \midrule
         &  &27.89 &0.829& 0.160 \\ 
        \checkmark &  & 28.32 & 0.837 & 0.156 \\
         & \checkmark & 28.52 & 0.843 & 0.153 \\
        \checkmark & \checkmark & 28.83 &0.849 &0.145 \\ 
        \bottomrule
    \end{tabular}}
\end{table}